\documentclass[letterpaper, 10 pt, conference]{ieeeconf}  
\pdfoutput=1

\IEEEoverridecommandlockouts                              

\usepackage[table,dvipsnames]{xcolor}
\usepackage{graphicx}
\usepackage{amsmath}
\usepackage{amssymb}
\usepackage{booktabs}
\usepackage{soul}
\usepackage{flushend}

\usepackage{subcaption}
\usepackage{graphicx}

\usepackage{arydshln}
\DeclareMathSymbol{\shortminus}{\mathbin}{AMSa}{"39}

\newcommand{\Du}{\mathcal{D}^{\mathrm{u}}}  
\newcommand{\Dl}{\mathcal{D}^{\ell}}        
\newcommand{\Xu}{{X}^{\mathrm{u}}}
\newcommand{\Xl}{{X}^{\ell}}

\newcommand{\Xut}{{X^\prime}^{\mathrm{u}}}
\newcommand{\Xlt}{{X^\prime}^{\ell}}

\newcommand{\Yl}{{Y}^{\ell}}
\newcommand{\Yp}{{Y}^{\mathrm{p}}}

\makeatletter
\let\NAT@parse\undefined
\makeatother
\usepackage[pagebackref,breaklinks,colorlinks,citecolor=blue]{hyperref}

\title{\LARGE \bf
T--UDA: Temporal Unsupervised Domain Adaptation\\in Sequential Point Clouds
}

\author{Awet Haileslassie Gebrehiwot$^{1}$, David Hurych$^{2}$, \\Karel Zimmermann$^{1}$, Patrick Pérez$^{2}$, Tomáš Svoboda$^{1}$ \\
\thanks{This work was supported in part by OP VVV MEYS funded project CZ.02.1.01/0.0/0.0/16 019/0000765 ``Research Center for Informatics'', by CTU Prague Project SGS22/111/OHK3/2T/13 and by Valeo. K. Zimmermann acknowledges CSF Project 20-29531S.}
\thanks{$^1$\,A. H. Gebrehiwot, K. Zimmermann and T. Svoboda are with the Faculty of Electrical Engineering, Czech Technical University in Prague}
 \thanks{$^2$\,D. Hurych and P. Pérez are with Valeo.ai}
}

\begin{document}

\maketitle

\begin{abstract}
Deep perception models have to reliably cope with an open-world setting of domain shifts induced by different geographic regions, sensor properties, mounting positions, and several other reasons. Since covering all domains with annotated data is technically intractable due to the endless possible variations, researchers focus on unsupervised domain adaptation (UDA) methods that adapt models trained on one (source) domain with annotations available to another (target) domain for which only unannotated data are available. Current predominant methods either leverage
semi-supervised approaches, e.g., teacher-student setup, or exploit privileged data, such as other sensor modalities or temporal data consistency.
We introduce a novel domain adaptation method that leverages the best of both trends. 
Our approach combines input data's temporal and cross-sensor geometric consistency with the mean teacher method.
Dubbed T-UDA for ``temporal UDA'', such a combination yields massive performance gains for the task of 3D semantic segmentation of driving scenes. Experiments are conducted on Waymo Open Dataset, nuScenes and SemanticKITTI, for two popular 3D point cloud architectures, Cylinder3D and MinkowskiNet.
Our codes are publicly available at \url{https://github.com/ctu-vras/T-UDA}.
\end{abstract}

\section{Introduction}
\label{sec:intro}
Autonomous vehicles and robots require a precise understanding of their dynamical environments to navigate around safely. In recent years we have witnessed the advent of powerful deep learning-based models to solve perception tasks such as 3D semantic segmentation with high-precision \cite{minkowskichoy20194d,hou2022point,cheng20212,yan20222dpass}. However, deep learning requires a massive amount of in-domain annotated data to reach such high precision and reliability. A common assumption for successful training of these perception models is that training and testing data share the same scene distributions, such as similar geographical regions, sensor placement and configuration, intensity channel features, etc.~\cite{triess2021survey}. However, in real-world applications, such as autonomous driving or robot navigation, this assumption does not hold because collecting and annotating massive datasets with 3D semantic labels for every new agent and domain is extremely expensive, time-consuming, and overall not scalable. Automotive companies have recently released several datasets~\cite{wod,nuscenes,semanticKitti}, with LiDAR recordings, but each with a different sensor configuration. These sensors work in different coordinate systems with different 3D sampling patterns, each covering distinct geographic regions and distributions of scene contents. As a result, deep learning-based perception models trained on one dataset do not adapt well to others, and the domain shift causes a considerable performance drop.
Each time a new LiDAR sensor configuration is selected, a new dataset has to be acquired and annotated to cover the new domain specifics. 

\begin{figure*}[t]
\begin{center}
    \includegraphics[width=0.85\textwidth]{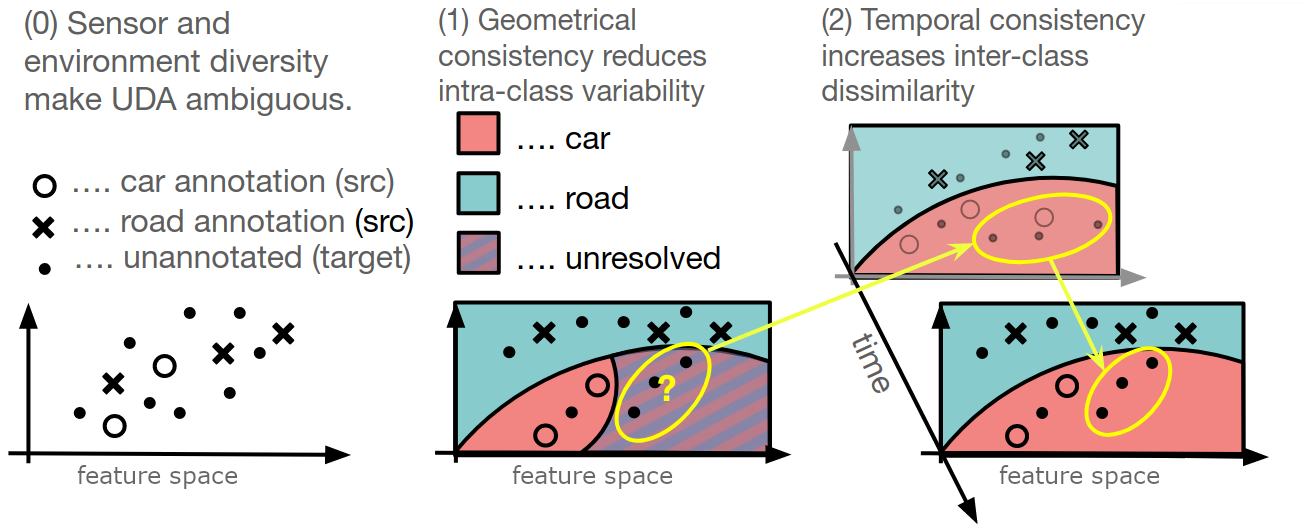}
\end{center}
    \caption{\textbf{Schematic illustration of the main ideas behind the proposed UDA approach.} (0) Huge intra-class variability caused by the sensor and environment diversity and inter-class similarity caused by low spatial resolution of distant objects make UDA challenging. (1) Enforcing a \emph{geometrically-consistent} representation reduces the cross-sensor variability and, hence, the intra-class variability, which in turn partially simplifies the underlying UDA; however, many unlabeled samples (highlighted by the yellow ellipse with `?'-mark) remain indistinguishable, causing difficulties to the a mean-teacher approach. 
    (2) \emph{Temporal consistency} constraints allow further reducing domain-shift ambiguities 
    by transferring the knowledge about the inter-class dissimilarity from the temporal neighborhood of the data;  The pair of pictures represents two scans of a scene at different times, showing how past scan points may help resolve current ones where fewer points cover the highlighted objects and become ambiguous.}\label{fig:overview}
\end{figure*}

We propose a novel two-stage architecture for unsupervised domain adaptation (UDA) that leverages the known 
geometrical and temporal consistency of measured data to reduce gaps, as illustrated in Fig.~\ref{fig:overview}. 
The overall architecture 
is shown in Fig.~\ref{fig:t-uda}, with its specific \emph{cross-sensor} geometric module detailed in Fig.~\ref{fig:cross-sensor}. 
While this module 
leverages LiDAR geometry to suppress the variability between the two sensors that can be explicitly modeled (sampling pattern, sensor position), the main architecture learns to suppress the other, difficult-to-model sources of distribution discrepancy, such as geographic location or object distribution. Consequently, the intra-class variability is reduced, and the inter-class dissimilarity is increased.

Contemporary domain adaptation methods often use domain transfer techniques~\cite{FerdinandDomainTransfer2020}, which solve the task by rendering semi-synthetic scans that match those of the target sensor sample patterns or use GAN-based domain adaption~\cite{jiang2021lidarnet} to transfer labels from a source to a target domain. One of the recent works~\cite{mean-teacher-segmentation} focuses on using teacher-student setup with averaging the network parameters to obtain supervision from a mean of all domain-consistent models -- usually referred to as \emph{the mean teacher}. On top of that, we also leverage the temporal consistency of the input data to constrain the mean of those models, which becomes consistent with the past and future measurements -- a \emph{temporally consistent mean teacher}.

Our contributions are three-fold. First and foremost, we identify a simple but important cross-sensor domain gap for LiDAR point clouds caused by LiDAR geometry, namely sampling differences and sensor positions. We propose reconstructing a similar cross-sensor representation from a sequence of point clouds to eliminate the discrepancies induced by the LiDAR configuration. Second, we present a novel multi-scan teacher-student learning framework for unsupervised domain adaptation of sequential point cloud data. Third, we provide thorough quantitative evaluations to validate our design choices on three datasets.
\section{Related Work}
\label{sec:relatedwork}

\noindent\textbf{3D Semantic Segmentation.~}
This paper focuses on domain adaptation for 3D semantic segmentation, one of the leading perception tasks. 3D semantic segmentation provides point-wise semantic labels for the 3D scene. Approaches for semantic segmentation may be categorized into (1) point-based, which directly operates on the three-dimensional points \cite{qi2017pointnet,qi2017pointnet++,thomas2019kpconv}, (2) projection-based, which operates on a different representation, like two-dimensional images \cite{wu2019squeezesegv2,lawin2017deep,Boulch2017UnstructuredPC} or three-dimensional voxel representations~\cite{yan20222dpass,zhu2021cylindrical,tang2020searching,minkowskichoy20194d}. Recent works, like Cylinder3D~\cite{zhu2021cylindrical} and MinkowskiNet~\cite{minkowskichoy20194d}, which are based on three-dimensional voxel representations, have shown a superior performance by aggregating multiple scans as input on various semantic segmentation benchmarks. Cylinder3D~\cite{zhu2021cylindrical} changes grid voxels to cylindrical in polar coordinates and applies an asymmetrical 3D convolution to boost performance. However, the network complexity and hardware requirements are restrictive for large-scale experiments. In contrast, the MinkowskiNet~\cite{minkowskichoy20194d} architecture adopts a voxel-based framework where sparse convolutions (SparseConv)~\cite{graham20183d} are utilized. Compared to traditional voxel-based methods, SparseConv only stores non-empty voxels and applies convolution operations only on these non-empty voxels more efficiently. In our experiments, we utilized both Cylinder3D and MinkowskiNet for 3D semantic segmentation of sequential point cloud data.

\smallskip\noindent\textbf{Unsupervised Domain Adaptation (UDA) for Point Cloud Segmentation.~}UDA for point cloud segmentation can be classified into two setups: (1) simulation-to-real~\cite{saltori2022cosmix,zhao2021epointda} and (2) real-to-real~\cite{yi2021complete,Ent,rochan2022unsupervised}. Simulation-to-real UDA is used when a deep learning-based model is trained with source domain from simulated or synthetically generated data and then tested on a target domain real-world data. In contrast, real-to-real UDA is used when a deep learning-based model is trained with source domain data of real-world scenes and then tested on target domain real-world data, often captured with a different LiDAR sensor and different geographic location. To this end, using images, Jaritz et al.~\cite{jaritz2020xmuda} proposed a framework to perform cross-modal domain adaptation for semantic segmentation. Ferdinand et al.~\cite{FerdinandDomainTransfer2020} proposed a domain transfer of a LiDAR-only semantic segmentation model, solving the task by rendering semi-synthetic scans that match the target sensor sample patterns. Jiang et al.~\cite{jiang2021lidarnet} propose to use GAN-based domain adaption to transfer labels from a source to a target domain. Yi et al.~\cite{yi2021complete} developed a 3D surface completion network to create a dense representation canonical domain from sparse point clouds of both source and target domain data and learn a segmentation network. In contrast to these methods, we use a simple geometric transformation and resampling of the point cloud of both source and target domains paired with \emph{time-aware mean Teacher model}, which guides the Student model training via pseudo-labeling of the target domain data.

\smallskip\noindent\textbf{Knowledge Distillation.~}
Knowledge distillation (KD) is an effective method of transferring knowledge from one model to another. This concept was first proposed in~\cite{model-compression}, where authors distilled a large model to a smaller one, allowing the latter to mimic the former, thus approaching its performance. Hinton et al.~\cite{KD} proposed an approach that enables a student network to learn the soft target output of a teacher network. Liu et al.~\cite{liu2018knowledge} distill an ensemble of teachers into one student. Tarvainen et al.~\cite{tarvainen2017mean} proposed a mean-teacher approach that averages model weights without changing the network architecture. Using the teacher network can also benefit from exploiting privileged information~\cite{chen2022automatic}. In our work, we adopt the mean teacher-based knowledge distillation framework, extending it to leverage a  temporal consistency of input data to constrain the mean on the teacher models in sequential scans for unsupervised domain adaptation.

\section{Method}
\label{sec:method}

\begin{figure*}[t]
    \begin{subfigure}{0.93\linewidth}
    \includegraphics[width=0.95\textwidth]{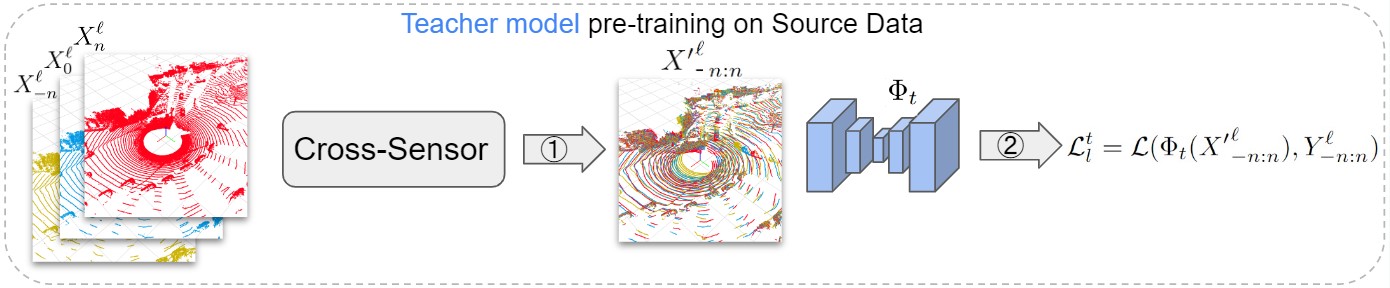}
    \caption{Teacher model pre-training on transformed source domain data.}
    \label{fig:teacher-training}
    \end{subfigure}
    \hfill
    \begin{subfigure}{0.99\linewidth}
    \includegraphics[width=0.99\textwidth]{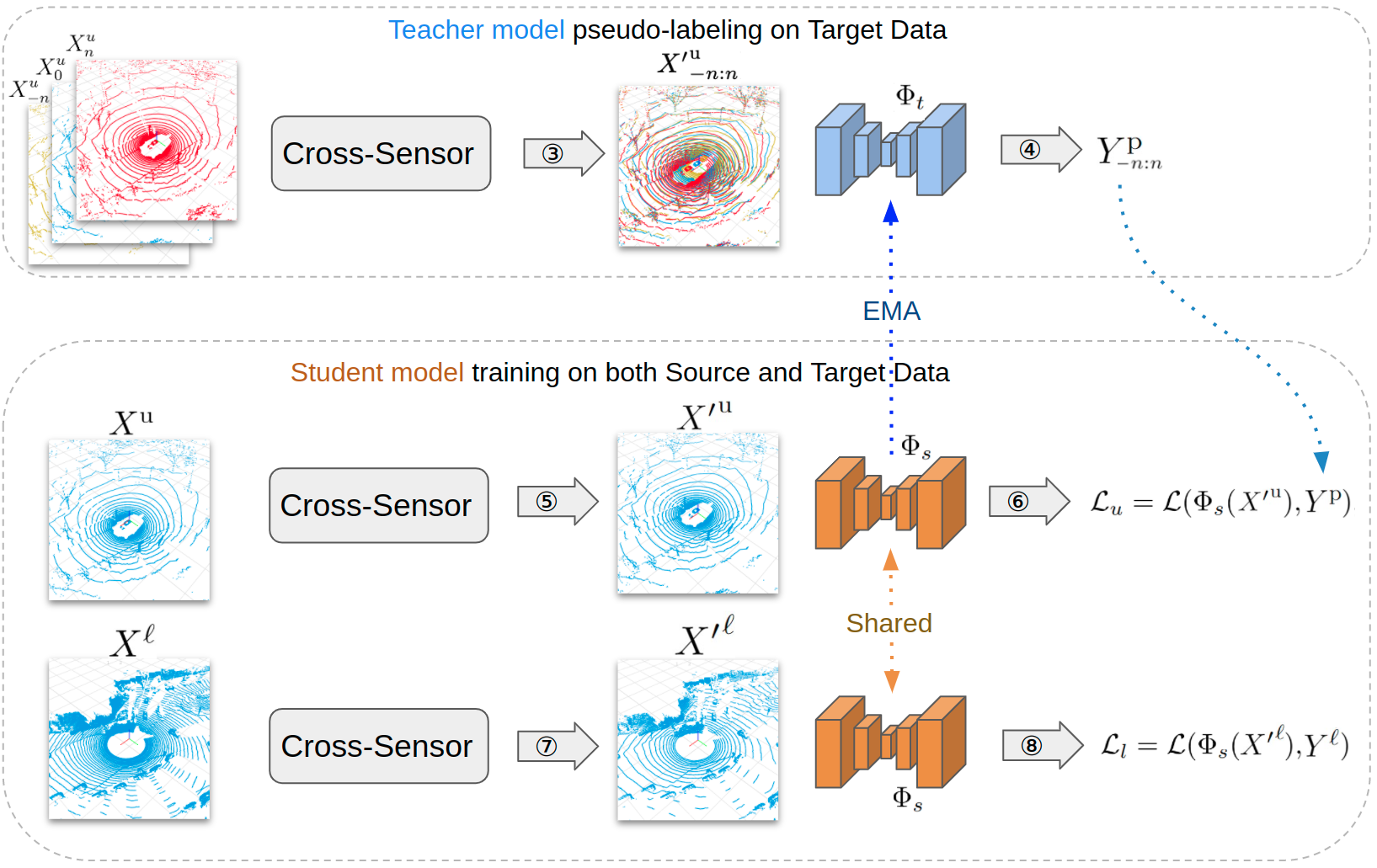}
    \caption{Student model training on transformed source and target domain data.}
    \label{fig:student-training}
    \end{subfigure}
    \caption{\textbf{Teacher-Student mutual training on the transformed source and target domain data}. Steps 1 and 3 apply our cross-sensor transformation to multiple individual scans and then merge the PC data from multiple time instances into one from the source or target domains separately. In step 2, ten epochs of warm-up training of the Teacher model are performed on the transformed and time-merged source domain data using labels. In step 4, the Teacher model infers pseudo-labels for the transformed target domain. In steps 5 and 7, the individual PC scans from the source and target domains are transformed using the cross-sensor module. In steps 6 and 8, the Student model is trained on a mixture of pseudo-labeled cross-sensor transformed target and labeled cross-sensor transformed source domain data, respectively.}
    \label{fig:t-uda}
\end{figure*}

We propose a temporal unsupervised domain adaptation framework to overcome the domain gap between scans captured by different LiDARs in different geographical regions. 
A point cloud (PC) $X = \{\mathbf{x}_p\}^P_{p=1}$ is a finite order-less collection of 3D points, where the number of points $P$ is assumed constant over time to keep the notation simple. We consider symmetric time-ordered PC sequences of form $X_{\shortminus n:n} = (X_{\shortminus n}, . . . , X_{\shortminus 1}, X_0, X_1, . . . , X_n)$ composed of a reference scan $X_0$, preceded by $n$ past scans and followed by $n$ future ones. All scans in the sequence are transformed into the coordinate system of the reference one. In standard UDA, we have a source domain with labeled LiDAR point clouds
$\Dl =\{(X^{\ell (i)},Y^{\ell (i)})\}^N_{i=1}$ 
and a target domain with unlabeled LiDAR point clouds $\Du = {\{X^{\text{u}(j)}\}^M_{j=1}}$, where $\Xl$ (resp. $\Xu$) denotes the labeled source (resp. unlabeled target) domain respectively and $\Yl$ denotes the semantic labels for the source domain;
$N$ and $M$ denote the number of point clouds in the source and target domains, respectively.
Observing that data from each domain have been collected with different LiDAR sensor configurations and from distinct geographic regions, we propose a two-stage approach. Fist, a cross-sensor data transfer bridges the domain gap arising from the different sampling patterns (assuming more scanning layers for the source-domain LiDAR) and sensor configurations. Second, a teacher-student learning technique is used to train a 3D semantic segmentation network on the labeled source $\Dl$ and unlabeled target $\Du$ domains to bridge the gap arising from the distinct geographic regions.

\subsection{Cross-Sensor adaptation stage}
\label{sec:method1}
\begin{figure*}[t]
    \centering
    \begin{subfigure}{0.95\linewidth}
    \includegraphics[width=0.99\textwidth]{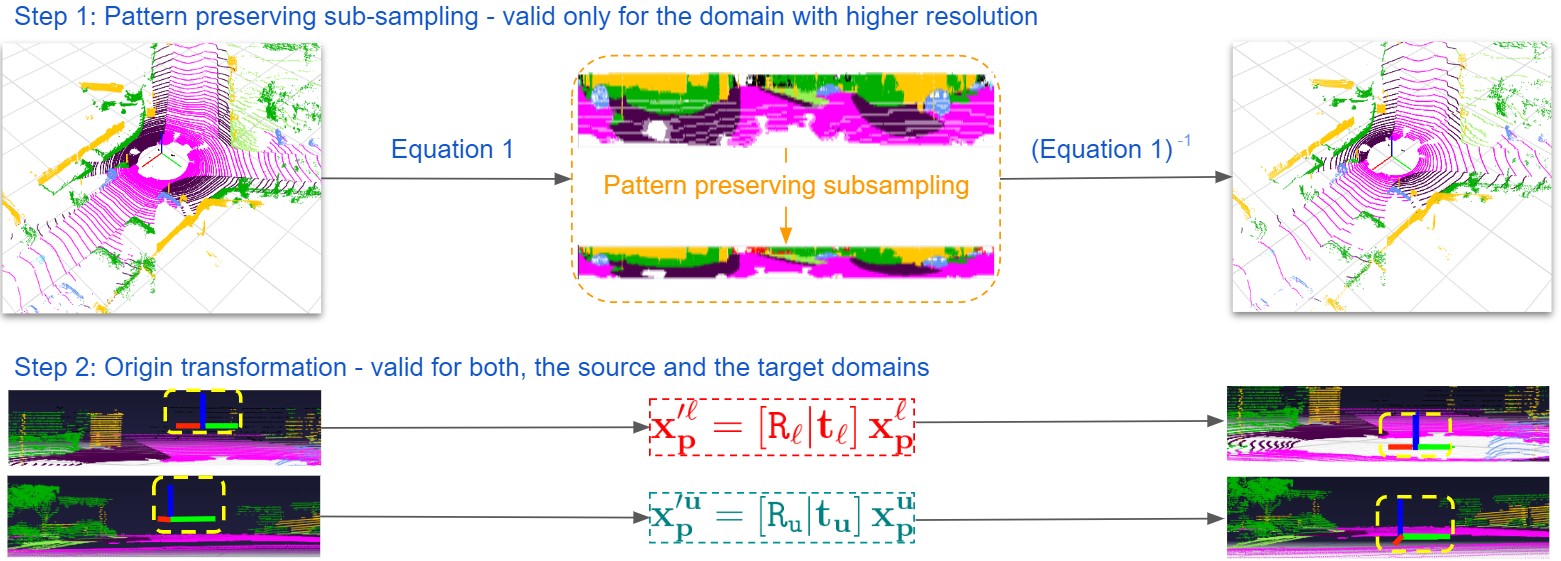}
    \end{subfigure}
    \caption{
    \textbf{Two-step cross-sensor LiDAR data transfer.} The first step (\textit{top}) is performed only for the domain with higher LiDAR resolution, while the second domain remains unchanged. Each separate scan is projected via Eq. (\ref{eq:2dprojection}) onto a spherical 2D representation, the number of points is then reduced, and the inverse projection is applied. The second step (\textit{bottom}) is applied to both the source and target domain data (one of them being already transformed by step 1). It consists of rotating and translating the data into the same coordinate system with the positions of the sensors placed on the ground plane.}
    
    \label{fig:cross-sensor}
\end{figure*}

This stage projects measurements from both source and target domains into a common space that minimizes cross-sensor variability yet preserves as much information as possible.
It consists of LiDAR sampling pattern transfer and LiDAR coordinate transformation (see Fig.\,\ref{fig:cross-sensor}).

\smallskip\noindent\textbf{LiDAR sampling pattern transfer.~} 
%
Since LiDAR sensors capture point clouds with a typical circular sampling pattern, it is essential to preserve this spatial arrangement and distribution in the common representation.
We follow prior works~\cite{jiang2021lidarnet, FerdinandDomainTransfer2020} and project the point clouds on their 2D spherical representation that allows for geometrically consistent subsampling. This projection transforms all 3D points $(x,y,z)$ into $W\times H$-dimensional 2D spherical image coordinates $(u,v)$ according to 
\begin{equation}
    \begin{array}{rcl}
   u &=& \frac{1}{2}[1-\arctan(y, x)\pi^{-1}]\,W  \\
   v &=& [1-(\arcsin(zr^{-1}) + f_{\text{up}})f^{-1}]\,H,
    \end{array}
\label{eq:2dprojection}
\end{equation}
where $r$ is the distance of the $(x,y,z)$-point from the center of projection and $(u,v)$ are the resulting 2D spherical coordinates, $f = |f_{\text{up}}| + |f_{\text{down}}|$ is the vertical field of view, $W$ and $H$ denote the width and the height of the LiDAR scan, and $\pi$ is the angle used to move the origin to the left edge of the 2D spherical image. In our experiments, the subsampling step works as follows: given source domain $\Xl$ with 64 laser beams and target domain $\Xu$ with 32 laser beams, we remove every 2nd laser beam from $\Xl$ to create a 32 beam source domain $\Xlt$ mimicking target $\Xu$. 

\smallskip\noindent\textbf{LiDAR coordinate transformation.~} 
Since there is no standard way of sensor mounting position, each time a new LiDAR sensor configuration is selected, the dataset is acquired with a different sensor setup. With the change in LiDAR mounting position, the acquired data statistics and distribution also change. This data distribution mismatch makes it unfeasible to apply the same model to different sensors in a naive way. Thus, point cloud segmentation models trained on data from one domain do not work well in another domain due to the misalignment of distribution. To overcome this limitation, we have incorporated a cross-sensor LiDAR coordinate transformation method to  project the source and target LiDAR coordinate system into a common coordinate system as shown in Fig.~\ref{fig:cross-sensor} step 2.

\subsection{Temporally consistent mean teacher domain adaptation}
\label{sec:method2}
In Fig.~\ref{fig:t-uda}, we show the block diagram of our proposed temporal unsupervised domain adaptation T-UDA. Let $(\Xlt, \Yl)$ be the transformed source domain point cloud with its semantic labels, where $\Xlt$ is the transformed source domain point cloud, and $\Yl$ is its point-level semantic class. Let the transformed target domain unlabelled point clouds be $\Xut$. In the teacher pseudo-labeling branch of Fig.~\ref{fig:t-uda}, a transformed multi-scan target point cloud $\Xut_{\shortminus n:n}$ 
is used to exploit the privileged information contained in the target domain sequential data. The output $\Yp$
is composed of the pseudo-labels that the teacher network produces during training. The \emph{student model training} branch involves self-training the student model using the target point cloud and pseudo-labels obtained from the teacher model, that is $(\Xut, \Yp)$, along with training the student model using the source point cloud and its corresponding labels, that is $(\Xlt, \Yl)$. Note that $\Xlt$ and $\Xut$ denote the outputs from our cross-sensor LiDAR transformation method. Lastly, let $\theta_{s}$ and $\theta_{t}$ be the student $\Phi_{s}$ and teacher $\Phi_{t}$ models' learnable parameters, respectively.

\paragraph{Teacher pre-training} 
Our T-UDA method aims to create a robust teacher model with access to privileged information in sequential point clouds and supervise the student model to learn from the target domain in a self-training manner. To this end, we created a teacher model $\Phi_t$ that can access both $n$ past and $n$ future scans, all transformed into the coordinate of the reference scan. The pre-training of the teacher model is accomplished by training it for a few epochs only on the source domain $(\Xlt_{\shortminus n:n}, \Yl_{\shortminus n:n})$ as a warm-up stage. After the warm-up stage, the teacher model switches from training to evaluation mode to generate pseudo-labels for the transformed target domain data $\Xut$ for the student model training. Since the network takes in aggregated multi-scan inputs $\Xlt_{\shortminus n:n}$ in the form of a point cloud with each point enriched by a time stamp $(\mathbf{x}_p,t)^{\top}$ and a semantic class label $C$, the network is supervised using the semantic class labels $\Yl_{\shortminus n:n}$ coming from all used times. During pre-training, the model is trained to minimize the loss 
\begin{equation}
\mathcal {L}_{\ell}^{t} = \mathcal{L}(\Phi_{t}(\Xlt_{\shortminus n:n}), \Yl_{\shortminus n:n})\,.
 \label{eq:st-loss-source}
\end{equation}
\paragraph{Student training}
Once the teacher model $\Phi_t$ pre-training stage is completed, we use the pre-trained teacher model parameters to initialize the student model $\Phi_s$. In contrast to the teacher model, the student does not use the past and the future scans at its input: it is trained purely on the reference 
scan. During training, we use the latest teacher model $\Phi_{t}$ to generate pseudo-labels $\Yp \in \Yp_{\shortminus n,n}$ for the unlabeled target domain $\Xut$, and we train the student $\Phi_{s}$ on pairs of $(\Xlt,\Yl)$ from the source and $(\Xut, \Yp)$ from the target domain. Thus, at each batch iteration the student parameters $\theta_{s}$ are updated to minimize the total loss 
\begin{equation}
\mathcal {L}_{\text{total}} = \mathcal{L}_{\ell} + \hat{C} \mathcal{L}_{\text{u}},
 \label{eq:loss-total}
\end{equation}
where $\hat{C}$ is the class probability of the pseudo label generated by the teacher model utilized to guide the training loss, $\mathcal{L}_{\ell}$ and $\mathcal{L}_{\text{u}}$ are the training loss functions of the student model on the source and target domain branches, respectively. Moreover, the individual segmentation losses of the student model are given by 
\begin{equation}
\mathcal {L}_{\ell} = \mathcal{L}(\Phi_{s}(\Xlt), \Yl), 
 \label{eq:loss-source}
\end{equation}
when trained on the source domain and 
\begin{equation}
\mathcal {L}_{\text{u}} = \mathcal {L}(\Phi_{s}(\Xut), \Yp), 
 \label{eq:loss-target}
\end{equation}
when trained on the target domain. 

The segmentation losses $\mathcal{L}_{\ell}$ and $\mathcal{L}_{\text{u}}$ are implemented as the standard cross-entropy loss. Lastly, we update the teacher parameters $\theta_{t}$ every iteration by the exponential moving average (EMA)~\cite{mean-teacher-segmentation}, i.e,
\begin{equation}
  \theta_{t} = \alpha \theta_{t} + (1-\alpha) \theta_{s}\,.
 \label{eq:ema}
\end{equation}

\section{Experiment}
\label{sec:exp}
We evaluate our method on a real-to-real domain adaptation setting on the three most popular autonomous driving datasets with two architectures. Since there is a lack of extensive research on real-to-real domain adaptation in 3D LiDAR point clouds, there is no standard baseline with a predefined set of object classes. Every method tried to create its new object categories by merging pre-existing object labels; Therefore, we compare our method with two state-of-the-art UDA methods, namely UDASSGA~\cite{rochan2022unsupervised}, complete\&Label (C\&L)~\cite{yi2021complete}  
and also with other methods such as (M+A)Ent~\cite{Ent}, SWD~\cite{SWD} and CORAL~\cite{coral} that were reported in the recent UDA paper UDASSGA~\cite{rochan2022unsupervised}. To make the comparison fair, we have divided the experiments into two sections according to the used subsets of classes on which the state-of-the-art methods are evaluated, see~\ref{sec:UDASSGA} and \ref{sec:completeandlabel}.

\subsection{Datasets}

We evaluate our approach on three widely used autonomous driving datasets, Waymo Open Data~\cite{wod}, nuScenes~\cite{nuscenes} and SemanticKITTI~\cite{semanticKitti}, each captured by a LiDAR with a different configuration at different geographic locations.

\smallskip\noindent\textbf{Waymo Open Dataset (WOD).} WOD contains LiDAR point cloud sequences from 1K scenes, each sequence containing about 200 scans. There are five LiDAR sensors.
We use the top 64-beam LiDAR in our experiments. The LiDAR scans are labeled with per-point semantic labels of 22 object categories, with one additional ``ignored'' class excluded from evaluations. The data is officially split into 798 training scenes and 202 validation scenes. Following the official recommendation, we use the 798 sequences for training and the rest 202 for evaluation.

\smallskip\noindent\textbf{nuScenes-lidarseg.} This dataset contains LiDAR point cloud sequences from 850 scenes, each containing about 40 scans annotated with per-point semantic labels. Different from Waymo Open, it is captured by a 32-beam LiDAR sensor with different configurations, causing a sampling gap from the Waymo Open point clouds. Officially these points are cast into 16 categories for the semantic segmentation task, with one additional “ignored” class excluded from evaluations. The dataset is officially split into 700 training scenes and 150 validation scenes. Following the official recommendation, we use the 700 sequences for training and the rest 150 sequences for evaluation.

\smallskip\noindent\textbf{SemanticKITTI.} This dataset was captured using Velodyne 64-beam LiDAR, similar to WOD, but with a different sensor configuration. It provides a large-scale set of driving-scene sequences for 3D semantic segmentation. Points are classified into 19 categories, with one additional “ignored” class excluded from evaluations. Following the official recommendation, we use sequences 00-07 and 09-10 for training and
sequence 08 for evaluation.

\subsection{Domain Adaptation Results}
This section presents the evaluation results of our proposed method for unsupervised domain adaptation (UDA) on point cloud semantic segmentation. The performance of our approach is examined and analyzed to demonstrate its effectiveness and applicability in real-world scenarios.
To this end, we have utilized the existing 3D semantic segmentation architectures that are able of processing multi-scan point cloud inputs, namely  MinkowskiNet~\cite{minkowskichoy20194d} and Cylinder3D~\cite{zhu2021cylindrical}. Please note that our \emph{main results are based on the MinkowskiNet} unless stated otherwise. The standard mean Intersection over Union (mIoU) metric is used for evaluation.

\subsection{T-UDA on SemanticKITTI (S) and nuScenes (N)}
\label{sec:UDASSGA}
For this experiment, we have used the SemanticKITTI and nuScenes datasets, first as a source domain $\Dl$ and the former as a target domain $\Du$ following the setup in~\cite{rochan2022unsupervised}. Since these datasets come with different amounts of labeled classes and different naming, we aligned them for domain adaptation be creating 11 intersecting semantic classes, namely \textit{car, bicycle, motorcycle, truck, other vehicle, pedestrian, drivable, sidewalk, terrain, vegetation and manmade} through merging and renaming some of the original semantic classes.

\begin{table*}[t]
\centering%
\caption{\textbf{nuScene to SemanticKITTI domain adaptation}. Comparison of 3D semantic segmentation performance (per-class mIoU and mean IoU) of proposed method with state-of-the-art UDA methods and no-adaptation (`No DA') baseline.}
\resizebox{0.8\textwidth}{!}{
\begin{tabular}{rrrrrrrrrrrrrrr}
    \toprule%
    & \rotatebox[origin=l]{90}{{\textbf{mIoU}}} & {\rotatebox[origin=l]{90}{car}} & {\rotatebox[origin=l]{90}{bicycle}} & {\rotatebox[origin=l]{90}{motorcycle}} & {\rotatebox[origin=l]{90}{truck}} & {\rotatebox[origin=l]{90}{o-vehicle}} & {\rotatebox[origin=l]{90}{person}} &  {\rotatebox[origin=l]{90}{drivable}} &  {\rotatebox[origin=l]{90}{sidewalk}} & {\rotatebox[origin=l]{90}{terrain}} & {\rotatebox[origin=l]{90}{vegetation}} & {\rotatebox[origin=l]{90}{manmade}} \\
    \midrule%
    No DA & 19.9 & 53.9 & 0.0 & 1.8 & 0.1 & 30.0 & 1.7 & 54.8 & 12.3 & 22.8 & 6.2 & 35.2 \\
    \hline
    SWD~\cite{SWD} &18.1&34.2&2.7&1.5&2.0&5.3&0.9&28.8&20.5&28.3&38.2&36.7& \\
    (M+A)Ent~\cite{Ent} &22.8&49.6&5.9&4.3&6.4&9.6&2.6&22.5&12.7&30.3&57.4&49.1& \\
    CORAL~\cite{coral} &23.3&47.3&10.4&6.9&5.1&10.8&0.7&24.8&13.8&31.7&58.8&45.5& \\
    UDASSGA~\cite{rochan2022unsupervised} &23.5&49.6&4.6&6.3&2.0&12.5&1.8&25.2&25.2&42.3&43.4&45.3& \\
    \rowcolor{gray!30}Ours & 49.0 & 93.0 & 0.0 & 11.4 & 3.4 & 47.0 & 15.7 & 83.3 & 54.4 & 67.9 & 83.9 & 79.4 \\
    \bottomrule%
\end{tabular}}%
\label{tab:nusc-kitti}%
\end{table*}%
\begin{table*}[t]
\centering%
\caption{\textbf{SemanticKITTI to nuScene domain adaptation}. Comparison of 3D semantic segmentation performance (per-class mIoU and mean IoU) of proposed method with state-of-the-art UDA methods and no-adaptation (`No DA') baseline.} 
\resizebox{0.8\textwidth}{!}{
\begin{tabular}{rrrrrrrrrrrrrr}
    \toprule%
    & \rotatebox[origin=l]{90}{{\textbf{mIoU}}} & {\rotatebox[origin=l]{90}{car}} & {\rotatebox[origin=l]{90}{bicycle}} & {\rotatebox[origin=l]{90}{motorcycle}} & {\rotatebox[origin=l]{90}{truck}} & {\rotatebox[origin=l]{90}{o-vehicle}} & {\rotatebox[origin=l]{90}{person}} &  {\rotatebox[origin=l]{90}{drivable}} &  {\rotatebox[origin=l]{90}{sidewalk}} & {\rotatebox[origin=l]{90}{terrain}} & {\rotatebox[origin=l]{90}{vegetation}} & {\rotatebox[origin=l]{90}{manmade}} \\
    \midrule%
    No DA & 21.0 & 49.1 & 0.2 & 0.8 & 1.0 & 0.2 & 8.7 & 44.5 & 16.9 & 16.7 & 54.6 & 38.2 \\
    \hline
    SWD~\cite{SWD}  &30.1&45.3&2.1&2.2&3.4&25.9&10.6&80.7&26.5&30.1&43.9&60.2&\\
    (M+A)Ent~\cite{Ent}  &32.0&57.3&1.1&2.3&6.8&23.4&7.9&83.5&32.6&31.8&43.3&62.3& \\
    CORAL~\cite{coral}  &33.3&51.0&0.9&6.0&4.0&25.9&29.9&82.6&27.1&27.0&55.3&56.7& \\    
    UDASSGA~\cite{rochan2022unsupervised} &34.5&54.4&3.0&1.9&7.6&27.7&15.8&82.2&29.6&34.0&57.9&65.7& \\
    \rowcolor{gray!30}Ours & 41.8 & 74.2 & 0.5 & 40.3 & 21.8 & 0.2 & 0.4 & 87.8 & 45.8 & 46.1 & 70.3 & 72.6 \\
    \bottomrule%
\end{tabular}}%
\label{tab:kitti-nusc}%
\end{table*}%
In this setup, we have two cases: (a) Domain adaptation from nuScenes to SemanticKITTI (N→S), where we use nuScenes as source domain $\Dl$ and SemanticKITTI as target domain $\Du$; (b) adaptation from SemanticKITTI to nuScenes (S→N) where  SemanticKITTI is the source domain and nuScenes the target one. Detailed results are reported in Tables~\ref{tab:nusc-kitti} and \ref{tab:kitti-nusc}.  As expected, testing on $\Du$ the network trained only on $\Dl$ without domain adaptation (`No DA') yields poor performance. With the proposed domain adaptation, our method achieves a new state-of-the-art result,  outperforming prior UDA methods by a considerable margin: When compared to the most recent method~UDASSGA~\cite{rochan2022unsupervised}, our method more than doubles the mIoU metric, exceeding UDASSGA by 25.5 mIoU on DA from nuScenes to SemanticKITTI. 

\subsection{Comparison to other UDA on all datasets combinations}
\label{sec:completeandlabel}

To understand the performance of our method against the state of the art on the three most popular autonomous driving datasets, we dedicated an experiment where we compare against the Complete\&Label (C\&L)~\cite{yi2021complete} on nuScenes~\cite{nuscenes}, SemanticKITTI~\cite{semanticKitti} and Waymo Open~\cite{wod} datasets. To make the comparison fair, we have followed the same approach as in \cite{yi2021complete}; to this end, we have used the same segmentation network architecture MinkowskiNet~\cite{minkowskichoy20194d} and we also evaluated on the same 10 classes, namely \textit{(car, bicycle, motorcycle, truck, o-vehicle, pedestrian, drivable, sidewalk, terrain, and vegetation)} for domain adaptation between SemanticKITTI and nuScenes, and on 2 classes \textit{(car, person)} for domain adaption between WOD and others as reported in~\cite{yi2021complete}.

As seen from Tab.~\ref{tab:uda-comparison}, our method significantly outperforms  Complete\&Label~\cite{yi2021complete} in all domain transfer setups. Specifically, our method achieved a relative gain of 7.1 and 12.3 mIoU on SemanticKITTI to nuScenes domain adaptation and vice-versa, respectively. Moreover, when we evaluate the domain adaptation method between WOD and nuScenes datasets, we observe a similar trend where the proposed method achieves a relative gain of 10.8, 2.3 mIoU on WOD to nuScenes and nuScenes to WOD, respectively. A similar trend is also observed when our method applied for domain adaptation between WOD and SemanticKITTI with a relative gain of 15.8, 11.3 mIoU on WOD to SemanticKITTI and SemanticKITTI to WOD, respectively. However, we have observed cases where the model trained on WOD as a source domain can work reasonably well even without any domain adaptation suggesting a good scene diversity in Waymo Open Dataset when compared to others.

\begin{table}[t]
    \centering
    \caption{\textbf{Additional comparisons}. LiDAR semantic segmentation performance of UDA between SemanticKITTI (K)~\cite{semanticKitti}, nuScenes (N)~\cite{nuscenes} and Waymo Open Dataset (W)~\cite{wod}. C\&L and Ours use the MinkowskiNet~\cite{minkowskichoy20194d}.}
    \setlength{\tabcolsep}{2.8pt}
    \begin{tabular}{rccccccc}
    \toprule
    method & K→N & N→K & W→N & N→W & K→W & W→K\\
    \toprule
    No DA & 19.3 & 18.4 & 48.2 & 38.2 & 29.7 & 69.4 \\ \midrule
    SqueezeSegv2~\cite{wu2019squeezesegv2} & 10.1 & 13.4 & 14.2 & 30.0 & 34.2 & 36.8 \\
    C\&L~\cite{yi2021complete} & 31.6 & 33.7 & 50.2 & 59.7 & 52.0 & 60.4\\
    \rowcolor{gray!30}Ours & 38.7 & 46.0 & 61.0 & 62.0 & 63.3 & 75.2\\
    \bottomrule
    \end{tabular}
    \label{tab:uda-comparison}
\end{table}

\begin{figure}[t]
  \centering
  \begin{subfigure}{0.49\linewidth}
    \includegraphics[width=0.99\textwidth]{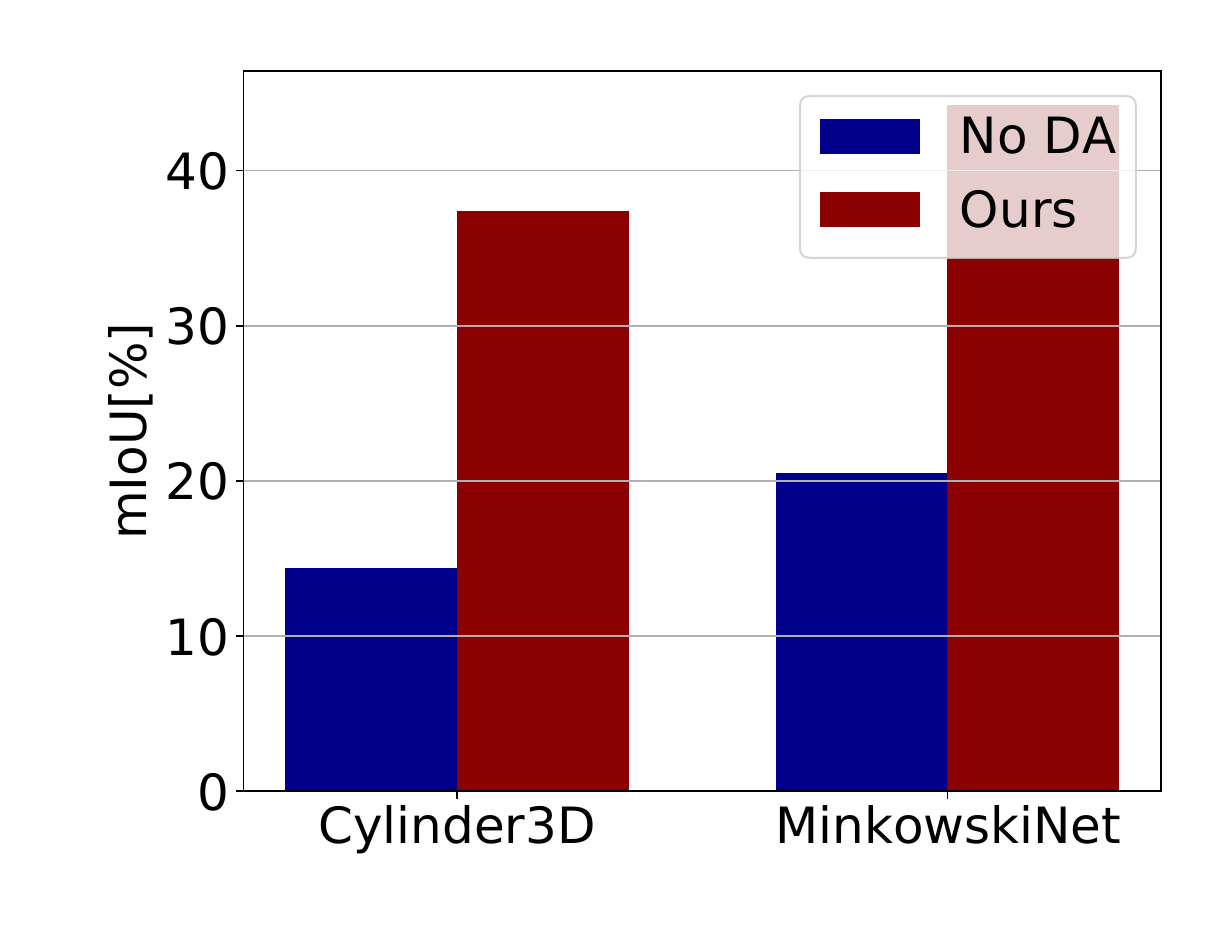}
    \caption{Generalization from K→N.}
    \label{fig:generalization-kn}
  \end{subfigure}
  \hfill
  \begin{subfigure}{0.49\linewidth}
  \includegraphics[width=0.99\textwidth]{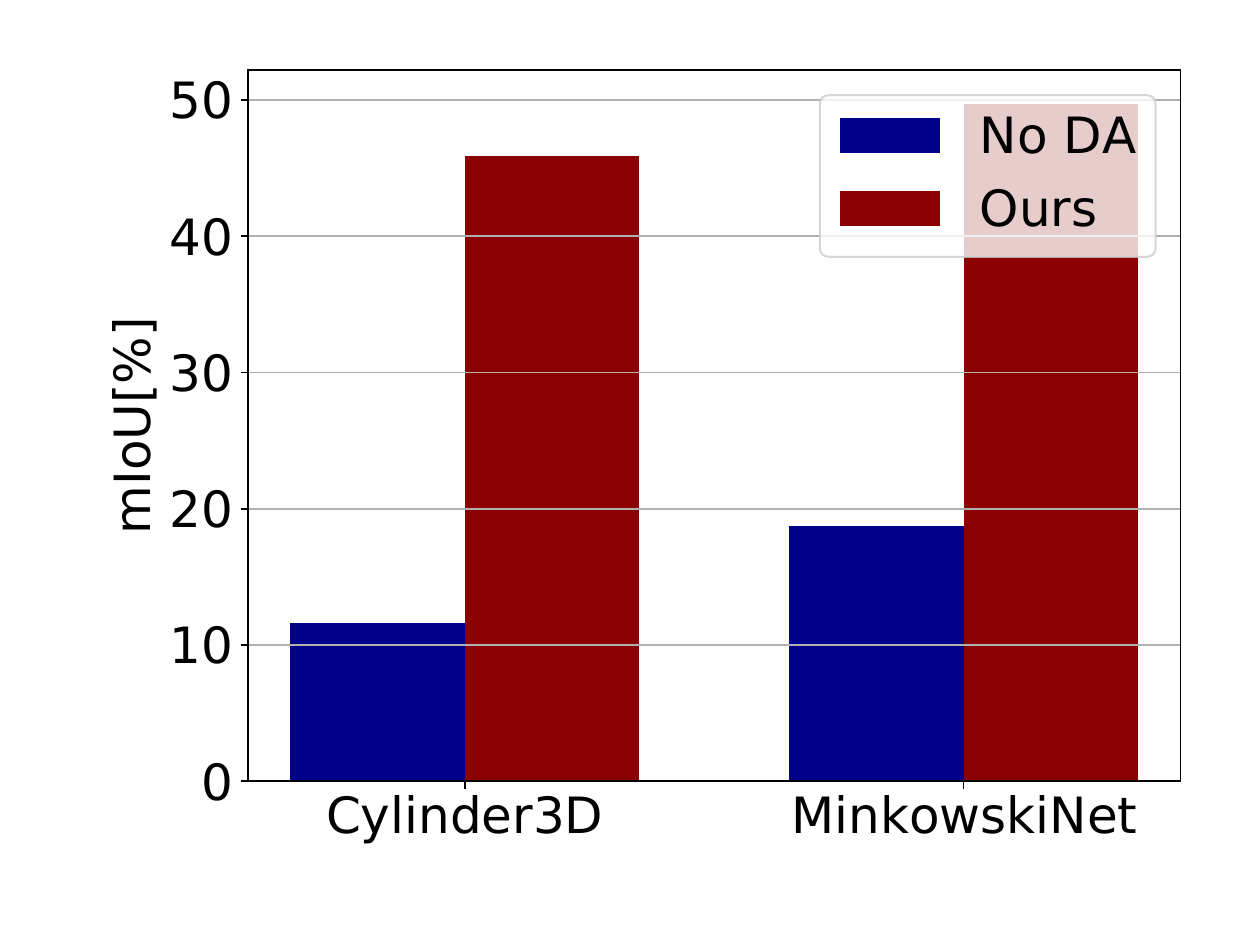}
    \caption{Generalization from N→K.}
    \label{fig:generalization-nk}
  \end{subfigure}
  \caption{\textbf{Generalization across architectures}. Capability of our method to handle various 3D segmentation architectures, namely Cylinder3D~\cite{zhu2021cylindrical} and MinkowskiNet~\cite{minkowskichoy20194d}.}
  \label{fig:generalization}
\end{figure}

\subsection{Ablation studies}
We examined our T-UDA components by categorizing them into four groups: LiDAR sampling pattern transfer (A), LiDAR coordinate transformation (B), temporally consistent mean teacher (C), and temporal (past and future) information (D). In Group A, we assessed the LiDAR sampling pattern transfer's importance in our cross-sensor adaptation stage 1 (\ref{sec:method1}). Similarly, in Group B, we evaluated the significance of the LiDAR coordinate transformation used in our cross-sensor adaptation stage 2 (\ref{sec:method1}). In Group C, we examine the importance of the temporally consistent mean teacher DA (\ref{sec:method2}). Lastly, in Group D, we explore the impact of incorporating temporal information, including past and future frames, into the teacher model. As shown in Tab. \ref{tab:my-label}, when applying only the LiDAR sampling pattern transfer DA (Group A) from SemantickITTI to nuScenes, and vice versa (K→N and N→K), there is a relative improvement of 9.1 and 6.8 mIoU, respectively, compared to the no-adaptation baseline (`No DA'). However, when the LiDAR coordinate transformation DA (Group B) is added to the LiDAR sampling pattern transfer DA, the relative improvement increases to 14.6 and 11.6 mIoU for K→N and N→K DA, respectively. Furthermore, incorporating the teacher-student mean teacher DA (Group C) provides an additional gain, resulting in 15.9 and 13.2 mIoU improvements over the baseline for K→N and N→K DA, respectively. Finally, adding temporal information to the temporally consistent mean teacher DA (Group D) results in further gains, leading to 21.1 and 30.0 mIoU improvements for K→N and N→K DA, respectively. The complete version of our T-UDA method, with all the components activated, achieves the best performance of 41.8 and 49.0 mIoU for K→N  and N→K, respectively.

To assess the generalizability of our approach, we have applied it to two different state-of-the-art architectures, namely MinkowskiNet~\cite{minkowskichoy20194d}, and Cylinder3D~\cite{zhu2021cylindrical}. As it can be seen from Fig.~\ref{fig:generalization}, our method brings very significant improvements to the no-adaptation baseline: 23.0 / 34.3 mIoU using Cylinder3D and 23.7 / 30.0 mIoU using MinkowskiNet when adapting from SemanticKITTI to nuScenes and vice-versa.  

\subsection{Implementation details} 
We used the same hyper-parameters as in MinkowskiNet and Cylinder3D, except that we add time as a fourth-dimensional feature. We set the batch size to 4 for semanticKITTI and nuScenes, and to 2 for WOD in both the source and target domains. The same proportion of samples from the source $(\Xlt,\Yl)$ and the target $(\Xut, \Yp)$ domains is used for student training. Our model is trained using stochastic gradient descent (SGD) with a learning rate of 0.24 and cosine learning rate decay for 40 epochs (10 for warmup and 30 for the student training) on an Nvidia A100 40GB GPU. 
For all experiments, we have used $n$ = 1, $\alpha=0.996$, and $\hat{C}$ as in \cite{t-concord3d}.

\subsection{Limitations}
An obvious limitation of our method is that it only allows subsampling data from a denser to a sparser LiDAR and not the other way around. A complementary approach could try to enrich the PC data of the domain with lower resolution and benefit from the richer features obtained for teacher-student training.

Our method in its current form does not allow handling domain adaptation for LiDARs with non-standard sampling/sweeping patterns, \emph{e.g.}, Livox vs. Velodyne LiDARs. As a part of our future work we plan to adapt the subsampling module to cope with arbitrary patterns.

\begin{table}[t]
    \centering
    \caption{\textbf{Effect of each component of T-UDA.} Ablation study of our LiDAR sampling pattern transfer (A), LiDAR coordinate transformation (B),  mean teacher domain adaptation (C) and temporal information (D). The importance of each component is evaluated on UDA between SemanticKITTI (K) and nuScenes
    (N), for the task of LiDAR 3D semantic segmentation (mIoU[\%]).}
    \begin{tabular}{ccccc|cc}
    \toprule
    {No DA}  & {A} & {B} & {C} & {D} & K→N & N→K \\
    \toprule
    \checkmark & & & & & 21.0 & 19.0\\
    &\checkmark & & &  & 30.1 &  25.8\\
    &\checkmark & \checkmark & & & 35.6 & 30.6 \\
    &\checkmark & \checkmark & \checkmark & & 36.9 & 32.2\\
    &\checkmark & \checkmark & \checkmark & \checkmark & \textbf{41.8} & \textbf{49.0}\\
    \bottomrule
    \end{tabular}
    \label{tab:my-label}
\end{table}

\section{Conclusion}
\label{sec:conc}
We presented a novel temporal unsupervised domain adaptation method, named T-UDA, that copes with differences in the number of laser beams, positions of sensors, and distributions of objects in LiDAR source and target domains. We have thoroughly tested our method on three publicly available and widely adopted datasets on two state-of-the-art deep neural network architectures trained for the 3D semantic segmentation task in point clouds. The performed ablation studies validate the contributions of the proposed modules. Our approach improves over the state-of-the-art methods by a large margin. On nuScenes to SemanticKITTI adaptation in particular, we improve the mIoU metric by 22\% over the so-far-best method \cite{yi2021complete} and we more than double the mIoU metric over the remaining state of the art, \emph{e.g.}, \cite{rochan2022unsupervised}.

{\small
\bibliographystyle{plain}
\bibliography{egbib}
}

\end{document}